\documentclass[10pt,twocolumn,letterpaper]{article}

\usepackage{iccv}
\usepackage{graphicx}
\usepackage{booktabs}
\usepackage{amsmath}
\usepackage{amsthm}
\usepackage{booktabs}
\usepackage{algorithm}
\usepackage{algorithmic}
\usepackage{epsfig}
\usepackage{amssymb}
\usepackage{bbold}
\usepackage{xcolor}
\usepackage{caption}
\usepackage{multirow}
\usepackage{soul}
\usepackage{bbm}
\usepackage{wrapfig}
\usepackage{comment}
\usepackage{enumitem}
\usepackage{arydshln}
\usepackage[title]{appendix}

\usepackage[pagebackref=true,breaklinks=true,letterpaper=true,colorlinks,bookmarks=false]{hyperref}

\iccvfinalcopy

\ificcvfinal\fi

\begin{document}

%%%%%%%%% TITLE
\title{AlignIT: Enhancing Prompt \underline{Align}ment in Custom\underline{i}za\underline{t}ion of Text-to-Image Models}

\author{Aishwarya Agarwal, Srikrishna Karanam, and Balaji Vasan Srinivasan\\
Adobe Research, Bengaluru India \\
{\tt \scalebox{.7}{\{aishagar,skaranam,balsrini\}@adobe.com}}
}

\twocolumn[{
\renewcommand\twocolumn[1][]{#1}
\maketitle
\begin{center}
 \centering
 \captionsetup{type=figure}
 \includegraphics[width=0.9\textwidth]{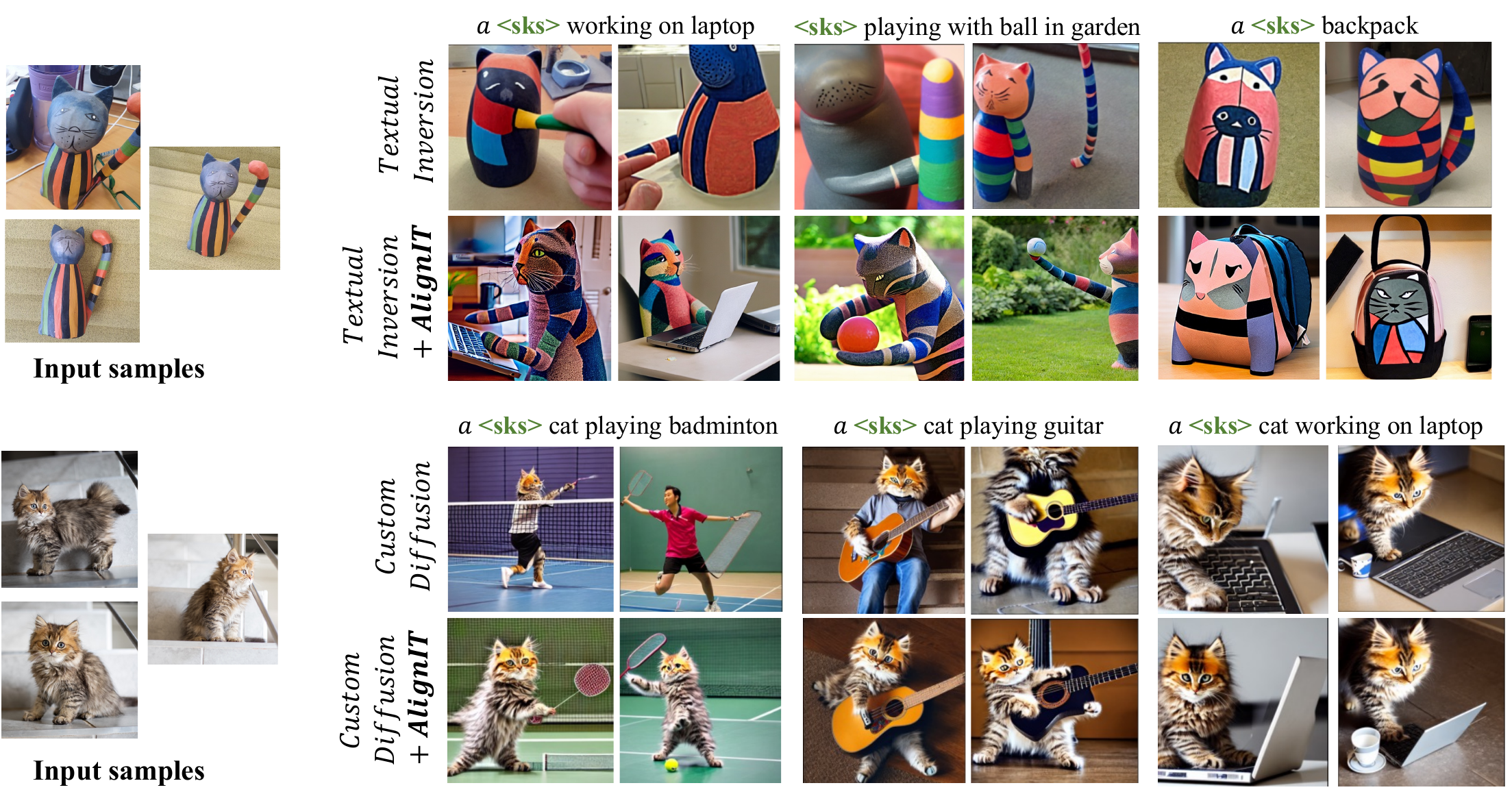}
 \vspace{-5pt}
 \caption{We propose a new algorithm, AlignIT, that can be used on top of any already-trained customization model to drastically improve the alignment of generated images with the text prompt directly at inference time, without requiring any retraining.}
 \label{fig:teaser_qual}
\end{center}
}]

\ificcvfinal\thispagestyle{empty}\fi

\begin{abstract}
 We consider the problem of customizing text-to-image diffusion models with user-supplied reference images. Given new prompts, the existing methods can capture the key concept from the reference images but fail to align the generated image with the prompt. In this work, we seek to address this key issue by proposing new methods that can easily be used in conjunction with existing customization methods that optimize the embeddings/weights at various intermediate stages of the text encoding process. 
 
The first contribution of this paper is a dissection of the various stages of the text encoding process leading up to the conditioning vector for text-to-image models. We take a holistic view of existing customization methods and notice that key and value outputs from this process differs substantially from their corresponding baseline (non-customized) models (e.g., baseline stable diffusion). While this difference does not impact the concept being customized, it leads to other parts of the generated image not being aligned with the prompt (see first row in Fig 1). Further, we also observe that these keys and values allow independent control various aspects of the final generation, enabling semantic manipulation of the output. Taken together, the features spanning these keys and values, serve as the basis for our next contribution where we fix the aforementioned issues with existing methods. We propose a new post-processing algorithm, \textbf{AlignIT}, that infuses the keys and values for the concept of interest while ensuring the keys and values for all other tokens in the input prompt are unchanged. 

Our proposed method can be plugged in directly to existing customization methods, leading to a substantial performance improvement in the alignment of the final result with the input prompt while retaining the customization quality. We conduct extensive experiments across various different customization methods and a wide variety of reference images and show consistent improvements both qualitatively and quantitatively.
\end{abstract}

\section{Introduction}
\label{intro}

We consider the problem of customizing the outputs of text-to-image diffusion models using concepts depicted in user-supplied reference images with a particular focus on improving the quality of alignment between input text prompts and the images generated using such customized models.

Building on top of the dramatic progress in text-to-image synthesis with text-guided diffusion models \cite{ho2020denoising,rombach2022high,sohl2015deep,song2020denoising}, there has been much recent work in customizing these models with user-supplied reference images \cite{gal2022image,kumari2023multi,han2023highly,zhao2023catversion}. Most of these methods embed knowledge from these reference images in the textual feature space as part of the text encoding process leading up to the conditioning vector used to condition the diffusion model. For instance,  \cite{gal2022image,han2023highly} introduced a new token into the text vocabulary (e.g. $<sks>$) and optimized its embedding to represent the custom concept of the reference image. On the other hand, CatVersion \cite{zhao2023catversion} proposed to optimize a select set of layers in the encoder model that produces the text feature vector, whereas Custom Diffusion \cite{kumari2023multi} tuned the parameters corresponding to the key and value weights. Across all methods, during inference, this embedded information of the custom concept is utilized as part of the text encoding process and new samples are synthesized.

From a usability perspective, it is crucial that customized text-to-image diffusion models are capable of generating images with custom concepts in novel scenes as described in the prompt. As noted in prior work \cite{gal2022image}, there are two key associated aspects. First, the model should be able to faithfully replicate the concept from the reference images (e.g., the cat from reference images should show up when that token is used in new prompts). Next, when using the custom token in a new prompt, the generated image must faithfully align the remaining parts of the scene with the prompt.

\begin{figure}[h]
    \centering
    \includegraphics[width = 0.80\linewidth]{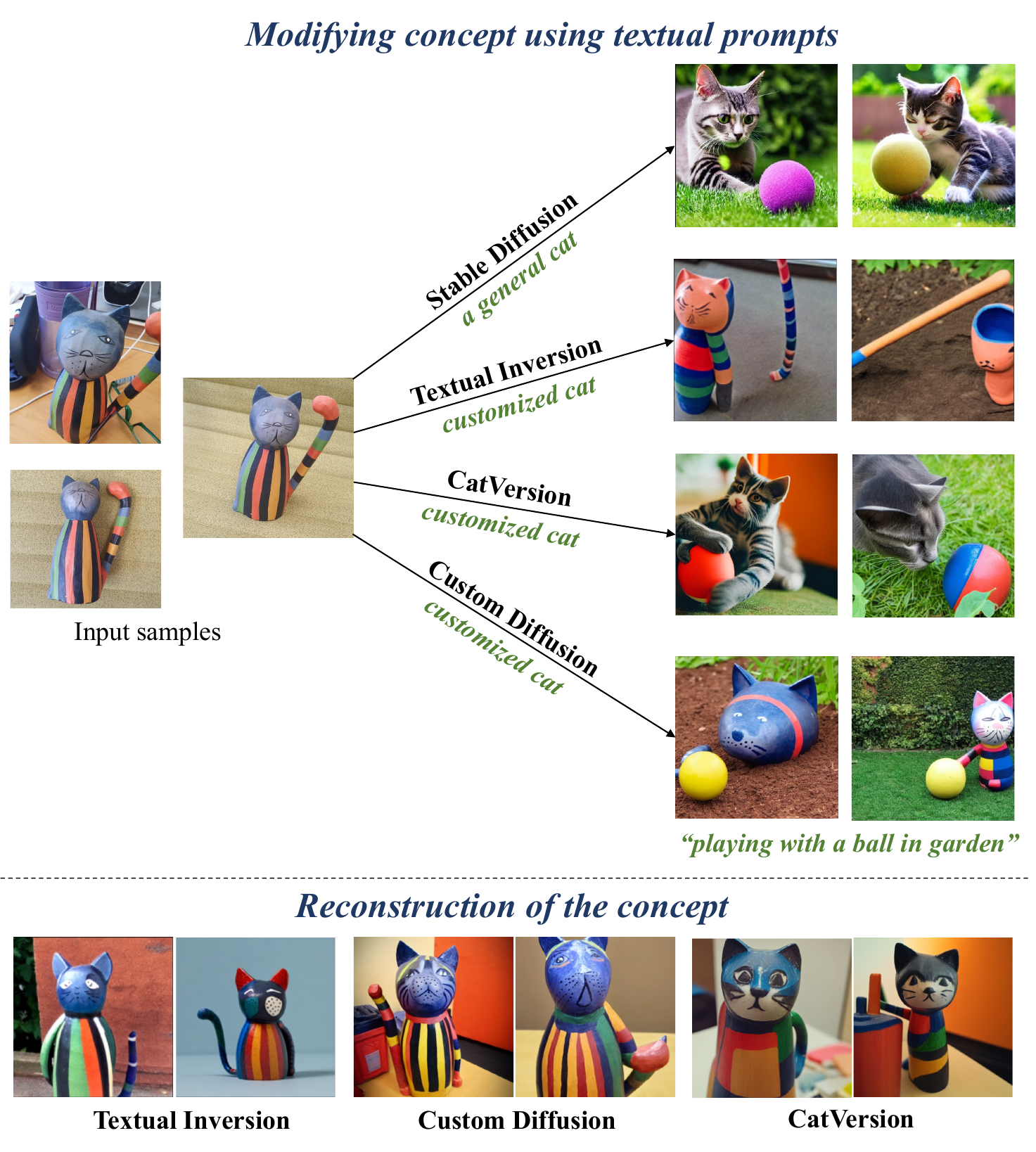}
    \vspace{-10pt}
    \caption{Editability-reconstruction tradeoff in baselines.}
    \label{fig:editability_lack}
\end{figure}

To understand this clearly, consider the example shown in Figure \ref{fig:editability_lack}. The first row shows images generated with s prompt \texttt{a cat playing with a ball in garden} using baseline stable diffusion. As can be seen from the outputs, the generated cat faithfully follows all aspects of the prompt (e.g., playing with ball, in a garden) suggesting the baseline model's good alignment between the image and the input prompt. The other three rows show images generated using Textual Inversion \cite{gal2022image}, CatVersion \cite{zhao2023catversion} and Custom Diffusion \cite{kumari2023multi} respectively, each model customized for the cat reference images with baseline stable diffusion. The results of these methods do not show the same level of alignment with the input prompt as the baseline results in the first row, suggesting a tradeoff between the reconstruction of the custom concept and the ability to edit this concept when used in novel generations. For example, in the second row, while the custom cat shows up in the first image, we do not see the \texttt{ball}. In the second image, while some aspects of \texttt{garden} starts showing up, it comes at the cost of deterioration in the custom cat. These results, part of a feature of such customization methods called reconstruction-editability tradeoff in \cite{gal2022image,tov2021designing,zhu2020improved}, show that these models are unable to allow control and modification of custom concepts as part of novel generations, limiting their practical usability.

To address the aforementioned limitations of existing customization techniques, this paper proposes a new algorithm, \textbf{AlignIT}, that can be easily plugged into these methods, immediately improving their performance. This comprises several contributions. First, we begin with a careful analysis of the text encoding process (from input prompt to output keys and values used for cross attention) involved in all these customization methods. While these methods adopt different strategies to optimize the text embedding, a common observation across all of them is that the keys and values they produce for a certain input prompt differs substantially from their corresponding baseline model. This difference, while helping produce custom concepts in the final result, also leads to other parts of the image not following the input prompt, leading to the misalignment between the output and the input text. Next, we also notice that these keys and values enable semantic manipulation and control over different aspects of the input prompt. 

\begin{figure}[t]
    \centering
    \includegraphics[width = \linewidth]{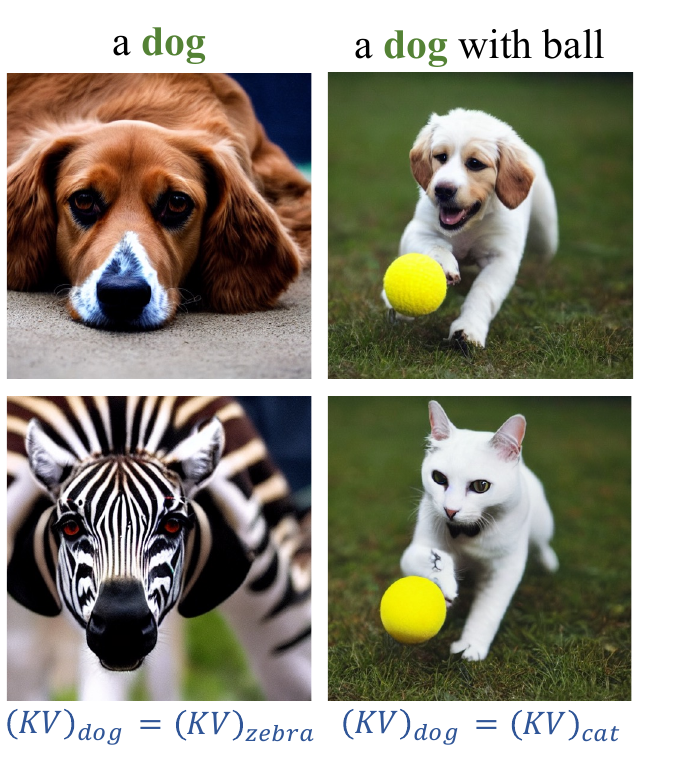}
    \vspace{-4pt}
    \caption{Control enabled by keys and values in cross-attention layers}
    \label{fig:kv_space_intro}
\end{figure}

For instance, in Figure~\ref{fig:kv_space_intro}, by replacing the keys and values (in every cross attention layer) of the \textit{dog} with a \textit{zebra}, we can generate a zebra even with \textit{a dog} as the input prompt in the first column. These two aspects (keys and values being different when compared to baseline stable diffusion and them being semantically manipulatable) are critical observations that motivate our next contribution, \textbf{AlignIT}, which when used with existing customization methods addresses their limitations.

Our key insight for AlignIT is to ensure only the keys and values corresponding to the custom concept of interest are modified in the text encoding process while keeping the keys and values for all other tokens in the input prompt unchanged from those in the custom model's corresponding baseline version (e.g., stable diffusion). 

As noted above, this is motivated by our observation that the customized model is able to reconstruct the custom concept using prompts like 

\texttt{a photo of a $<$sks$>$} (see Figure 2 bottom). This suggests the keys and values of the learned embedding during customization has all the information about the concept from the reference images and we just have to ensure keys and values for other parts of the prompt (which also change with existing customization methods) do not change. Given a model trained with one of the above customization methods, we achieve this with a test-time-only adaptation by using the keys and values of the object from those computed with the custom model. Since our method involves the test-time manipulation of keys and values, it easily can be used in conjunction with any customization approach that optimizes text embedding during its training process. We conduct extensive experiments using the CustomConcept101 dataset and demonstrate our approach substantially improves the customization capabilities of three different existing methods. In Figure~\ref{fig:teaser_qual}, we show some sample results. In the first row, with AlignIT, we are able to improve the quality of the textual inversion model (e.g., in the first column, baseline textual inversion does not generate a laptop which our method is able to correct). Similarly, in the second row, we show improved results with Custom Diffusion (e.g., in the first column, our method not only depicts badminton but also faithfully reconstructs the cat).

To summarize, our key contributions in this work are:
\begin{itemize}
    \item We dissect various stages of the text encoding process and discuss reasons why existing customization methods fail to generate images fully aligned with text prompt (as shown in Figure \ref{fig:editability_lack}). We notice that key and value outputs from the text encoding process differ substantially from their corresponding non-customized models (e.g., baseline stable diffusion).
    \item We demonstrate that the keys and values allow independent control over various aspects specified as part of the text prompt, enabling semantic manipulation of the generated image as shown in Figure \ref{fig:kv_space_intro}.
    \item We propose a novel algorithm, AlignIT, that utilises the identified properties of the keys and values and substantially improves the alignment of generated images with the prompt. AlignIT is a training-free algorithm that can be plugged into an already-trained customization model to improve its performance directly during inference.

\end{itemize}

\section{Related Work}

With remarkable advancements in text-to-image synthesis with text-guided diffusion models \cite{ho2020denoising,rombach2022high,sohl2015deep,song2020denoising}, there has been much recent work in customizing these models with user-supplied reference images \cite{gal2022image,agarwal2023image,han2023highly,kumari2023multi,zhao2023catversion,tewel2023key,han2023svdiff,ruiz2023dreambooth,gal2023encoder,meng2022locating}.

Many customization methods embed knowledge from these reference images in the textual feature space as part of the text encoding process leading up to the conditioning vector used to condition the diffusion model. \cite{gal2022image,han2023highly} introduce a new token into the text vocabulary and optimized its embedding to represent the custom concept of interest (while keeping diffusion model weights fixed), which could then be used to synthesize novel customized variations of the reference image. CatVersion \cite{zhao2023catversion} finetunes the weights of the attention layers in the text encoder. CustomDiffusion \cite{kumari2023multi} finetunes only the key and values weights in the cross-attention layers of the UNet to invert the concept of interest into a rare token. Han et al. \cite{han2023svdiff} uses singular value decomposition to finetune the singular value matrix of the diffusion model backbone, significantly reducing the number of parameters needed for learning the target concepts. Perfusion \cite{tewel2023key} as well updates the key and value weights while introducing a gated Rank-one Model Editing \cite{meng2022locating} to make it easier to combine multiple concepts.

These existing methods though are able to reconstruct the custom concept of interest, but they often struggle to generate images that align fully with the text prompt. Prior works on aligning image and text have tackled this through attention-map re-weighing \cite{feng2022training,phung2023grounded,wu2023harnessing}, latent-optimization \cite{chefer2023attend,agarwal2023star,rassin2023linguistic}, but none of these method address the alignment issue of customization methods and they instead aim to enhance the base models in generating text-aligned images. In this work, we seek to address this key issue by proposing new methods that can easily be used in conjunction with existing customization methods that optimize the embeddings/weights at various intermediate stages of the text encoding process. 

\section{Approach}

As noted in Section \ref{intro}, existing customization techniques can represent the custom concept of interest but when used to generate this concept in new scenarios, the outputs do not accurately align with the user's intent/input prompt. 

One key observation from Section 1 was the keys and values these techniques produce differ substantially from those produced by the corresponding baseline model, e.g., stable diffusion. To understand this better and how it informs our proposed method, we first begin with a brief summary of the text encoding process in text-to-image diffusion models, followed by a discussion on why existing methods fail and our proposed solution to address these issues. 

\subsection{Text encoding process}
\begin{figure}[t]
    \centering
    \includegraphics[width = 0.8\linewidth]{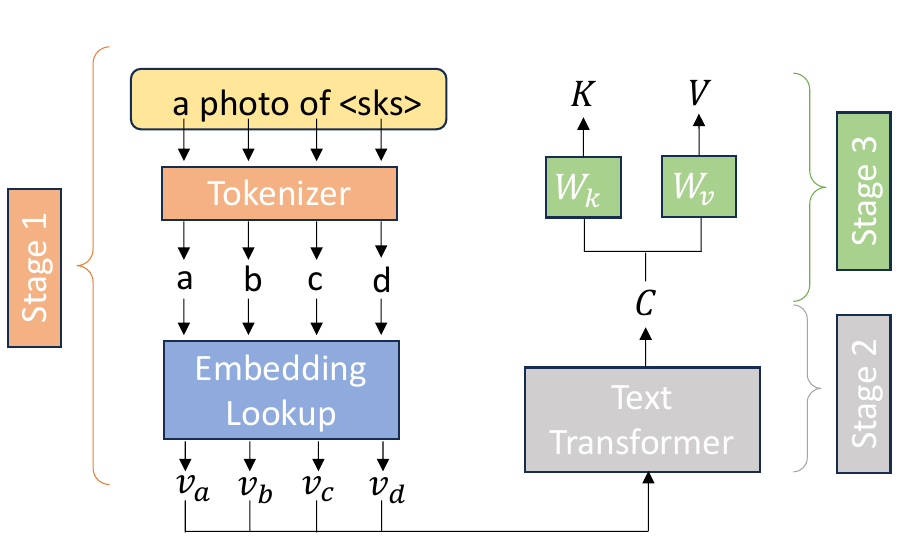}
    \vspace{-4pt}
    \caption{Various stages of the text encoding process.}
    \label{fig:text_encoding_stages}
\end{figure}

Given a text prompt (e.g., \texttt{a photo of $<sks>$}), there are several steps involved in producing the conditioning vector that feeds into the cross-attention layers of the noise prediction model (see Figure \ref{fig:text_encoding_stages}):\\
\textbf{Stage 1.} Each word/subword in the input prompt is tokenized, and each token's embedding is retrieved from a precomputed database. This constitutes the first stage of the encoding process. Customization techniques such as \cite{gal2022image,han2023highly} use this stage to optimize the model and infuse custom-concept knowledge. For instance, using a placeholder string (e.g., $<sks>$ in Figure \ref{fig:text_encoding_stages}), \cite{gal2022image} optimizes a reconstruction loss objective and learns a new embedding (e.g., $v_d$ for $<sks>$) for this placeholder string, essentially augmenting the existing vocabulary with this new information. During inference, given a new prompt with the $<sks>$ string, this new vocabulary is used to compute the conditioning vector and generate the output. \\
\textbf{Stage 2.} Given the per-token embeddings from stage 1, the next step is to compute the final text encoding $C$ using the diffusion model's text encoder (e.g., CLIP). Unlike \cite{gal2022image} or \cite{han2023highly}, CatVersion \cite{zhao2023catversion} uses the last three attention layers within this text encoder module to embed knowledge of the custom concept (i.e., they modify the weights of these three attention layers). \\
\textbf{Stage 3.} The vector $C$ from stage 2 is then input to all cross attention layers of the diffusion model's noise prediction module (e.g., UNet \cite{ronneberger2015u} in stable diffusion). Each cross-attention layer's key and value matrices, $W_k$ and $W_v$ respectively, are used to project $C$ into output keys ($K$) and values ($V$). Different from \cite{gal2022image,zhao2023catversion}, another line of customization methods \cite{kumari2023multi} tune these $W_k$ and $W_v$ matrices to embed the custom-concept knowledge.

With our proposed method, discussed below, we seek to improve the performance of existing customization methods that optimize embeddings or weights during any of the stages in the text encoding process described above.

\begin{figure*}[h]
    \centering
    \includegraphics[width = 0.95\linewidth]{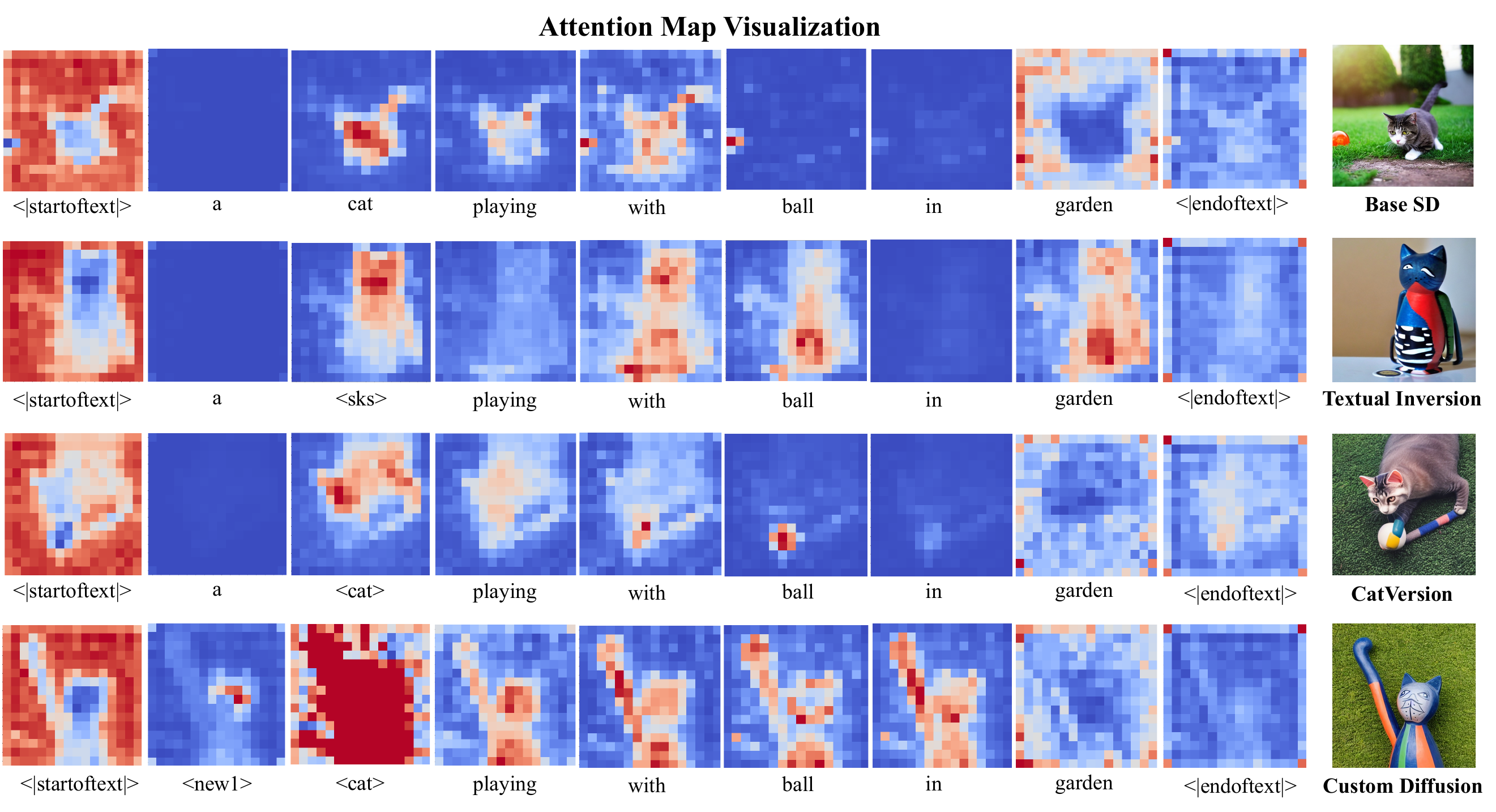}
    \caption{Cross-attention maps to demonstrate that baselines undesirably impact keys/values of tokens other than the concept of interest too.}
    \label{fig:ca_maps}
\end{figure*}

\subsection{Why do existing customization methods fail?}
As shown in the discussion above, existing customization methods embed the custom knowledge from reference images at one of the three stages of the text encoding process. During inference, this is used (either learned embeddings as in \cite{gal2022image} or text encoder attention weights \cite{zhao2023catversion} or cross-attention weight matrices \cite{kumari2023multi}) to generate the keys and values given the input prompt, which then are used to perform denoising. Since any of these three types of optimization eventually lead to the output keys ($K$) and values ($V$) from stage 3 (in green in Figure~\ref{fig:text_encoding_stages}), these $K$ and $V$ matrices are the only factors that influence/control the impact of the input text prompt on the final generated image. This means, in the context of existing customization methods, the quality of output depends on how well the information gets propagated from the input prompt to the final $K$ and $V$ matrices from stage 3. As shown in Figure~\ref{fig:editability_lack}, while this helps these methods reconstruct the custom concept of interest (e.g., the cat), they are unable to accurately generate other parts of the scene described in the input prompt.

To understand why this is the case, let us first begin with their corresponding baseline model (pretrained stable diffusion). Consider the first row in Figure~\ref{fig:ca_maps} where we see an image generated by this baseline model for \texttt{a cat playing with a ball in garden}. The attention maps show all the key attributes in the prompt, \texttt{cat}, \texttt{ball} and \texttt{garden}, are well represented, suggesting the $K$ and $V$ outputs from stage 3 for each of the tokens (e.g. cat, ball etc) has all the required information properly propagated from the input text prompt via the three encoding stages of Figure \ref{fig:text_encoding_stages}.

Next, consider the result with textual inversion \cite{gal2022image} in row 2 in Figure~\ref{fig:ca_maps}. Here, one can note while the custom concept's attention map (corresponding to $<sks>$) is well highlighted, the attention maps for the \texttt{ball} and \texttt{garden} tokens differ substantially from what the baseline model produced, resulting in an undesirable output. This difference is because the custom concept's optimized embedding $v_{<sks>}$ (note that the corresponding baseline model's embedding for this token is $v_{d}$ in Figure \ref{fig:text_encoding_stages})  ends up impacting the keys and values of tokens other than $<sks>$ as well. This happens since the $v_{<sks>}$ vector (along with $v_{a}$, $v_{b}$, and $v_{c}$) produces an input to the text transformer in Figure \ref{fig:text_encoding_stages} that is different from the baseline (since $v_{d}$ is now different from $v_{<sks>}$), resulting in a $C$ that is different from the baseline. A similar phenomenon can be noticed with CatVersion and custom diffusion methods as well (see rows 3 and 4 in Figure~\ref{fig:ca_maps}). Whereas CatVersion optimizes the text transformer directly, custom diffusion modifies the $W_k$ and $W_v$ values of Figure \ref{fig:text_encoding_stages}. These modifications end up impacting all the tokens, leading to either the custom cat not showing up or the ball/garden missing from the output.

The aforementioned observations motivate our proposed method, \textbf{AlignIT}, that allows more controlled infusion of custom knowledge into the model while not impacting the keys and values of other tokens in the input prompt, leading to both the retention of the custom concept as well as better alignment of the final generation with the input prompt.

\subsection{AlignIT}
Based on the observations from the previous section, the key insight for AlignIT is that the keys and values for any customization method should differ (when compared to the baseline model, e.g., stable diffusion) only for the custom token. This way, we can ensure both reconstruction of the custom concept and also adhere to the input prompt as closely as possible. This idea leads to our proposed method which we show can be used with any of the existing customization methods discussed previously and improve their performance both qualitatively and quantitatively.

\begin{figure}[t]
    \centering
    \includegraphics[width = \linewidth]{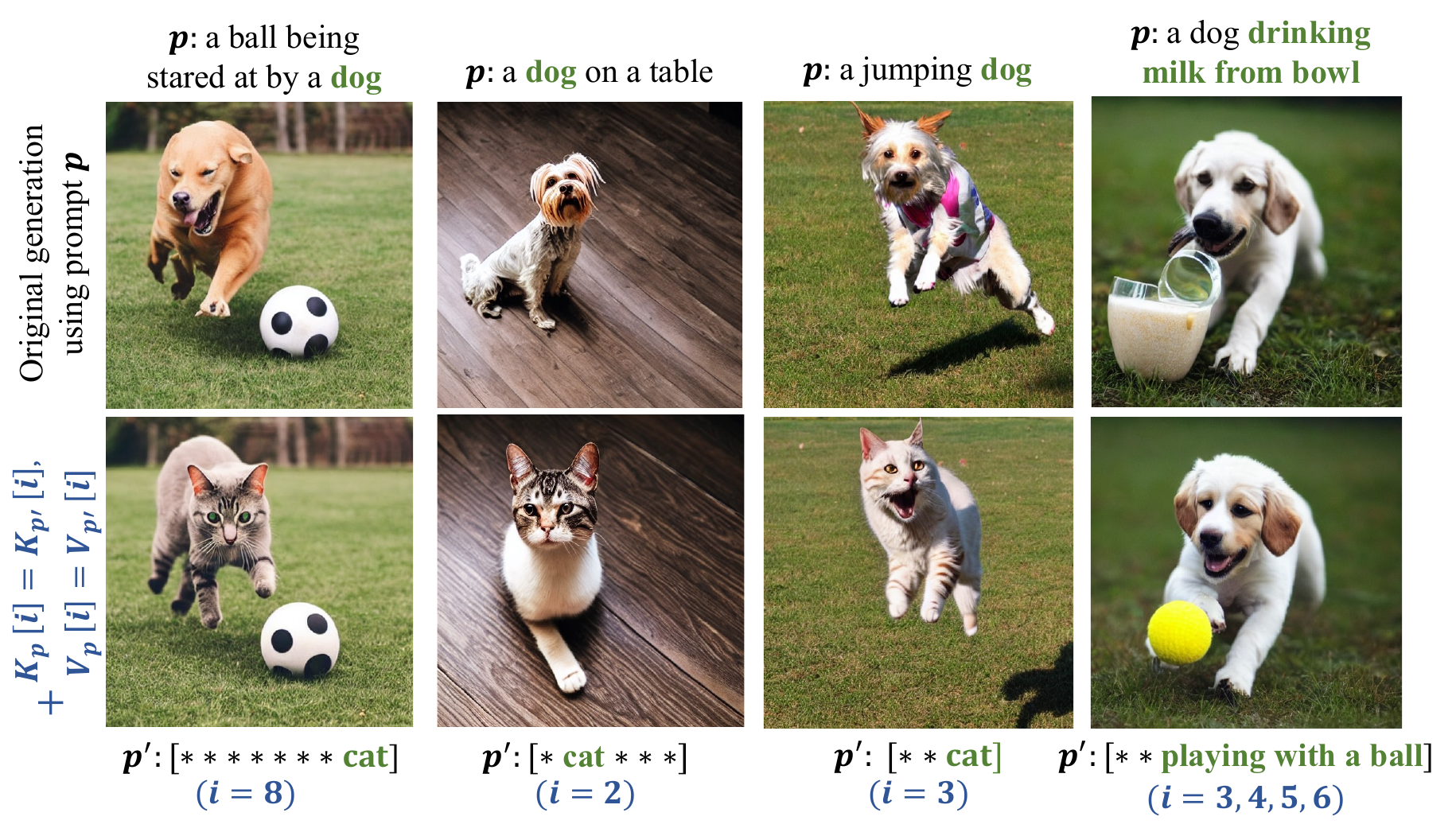}
    \vspace{-8pt}
    \caption{Semantic manipulations offered by keys and values.}
    \label{fig:kv_space_analysis1}
\end{figure}

Before discussing the details of AlignIT, we first explain how these keys and values allow for manipulation and independent control over various aspects of the input text prompt. This will then lead to the main ideas of our method. Let us begin with an example. Consider a prompt $p$ \texttt{a jumping dog} (see third column in Figure \ref{fig:kv_space_analysis1}). The first row in the third column shows the result with the baseline stable diffusion model. To generate the image in the second row, we follow the steps below: 

\begin{itemize}
    \item We keep the input prompt $p$ (\texttt{a jumping dog}) unchanged
    and compute the $K_p$ $\in$ $\mathcal{R}^{n\times d}$ and $V_p$ $\in$ $\mathcal{R}^{n\times d}$ for each cross-attention layer ($n$ is the number of tokens and $d$ is the feature dimensionality of the layer) as part of Stage 3 in Figure~\ref{fig:text_encoding_stages}.
    \item Given a new concept $o$ (\texttt{cat} here) with which we want to replace/edit the main object in $p$ (i.e., \texttt{dog}), we first construct a dummy prompt $p'$ such that $p'$ has the token $o$ at the same index $i$ ($3$ in this case) as the input prompt $p$. We pad the remaining token slots with placeholders (e.g. $*$) that have no significance and do not impact the generation.
    \item We next use the dummy prompt $p'$ and compute the keys $K_{p'}$ and values $V_{p'}$ for all the cross attention layers. Since the only token carrying significance in prompt $p'$ is our concept of interest $o$ at index $i$, $K_{p'}[i]$ and $V_{p'}[i]$ have all the knowledge required for capturing concept $o$ in the final generation. 
    \item Before feeding the $K_p$ and $V_p$ computed in step 1 above for further steps of denoising, we modify them as follows (in each timestep):
    \begin{equation}
        K_p[i] = K_{p'}[i], V_p[i] = V_{p'}[i]
    \end{equation}
    In other words, we copy the keys and values of this new concept (\texttt{cat}) into the location that previously had \texttt{dog}'s keys and values, while keeping all other keys and values unchanged. This way we end up generating images that follow all aspects of the original prompt $p$ (e.g. \texttt{jumping}) while replacing the semantic concept (\texttt{dog} here) with the concept of interest (\texttt{cat} in this case). This is indeed the case in the third column/second row of Figure~\ref{fig:kv_space_analysis1} where the dog is replaced by cat, while retaining other semantics (\texttt{jumping}) from the original image.
\end{itemize}

\begin{figure}[h]
    \centering
    \includegraphics[width = \linewidth]{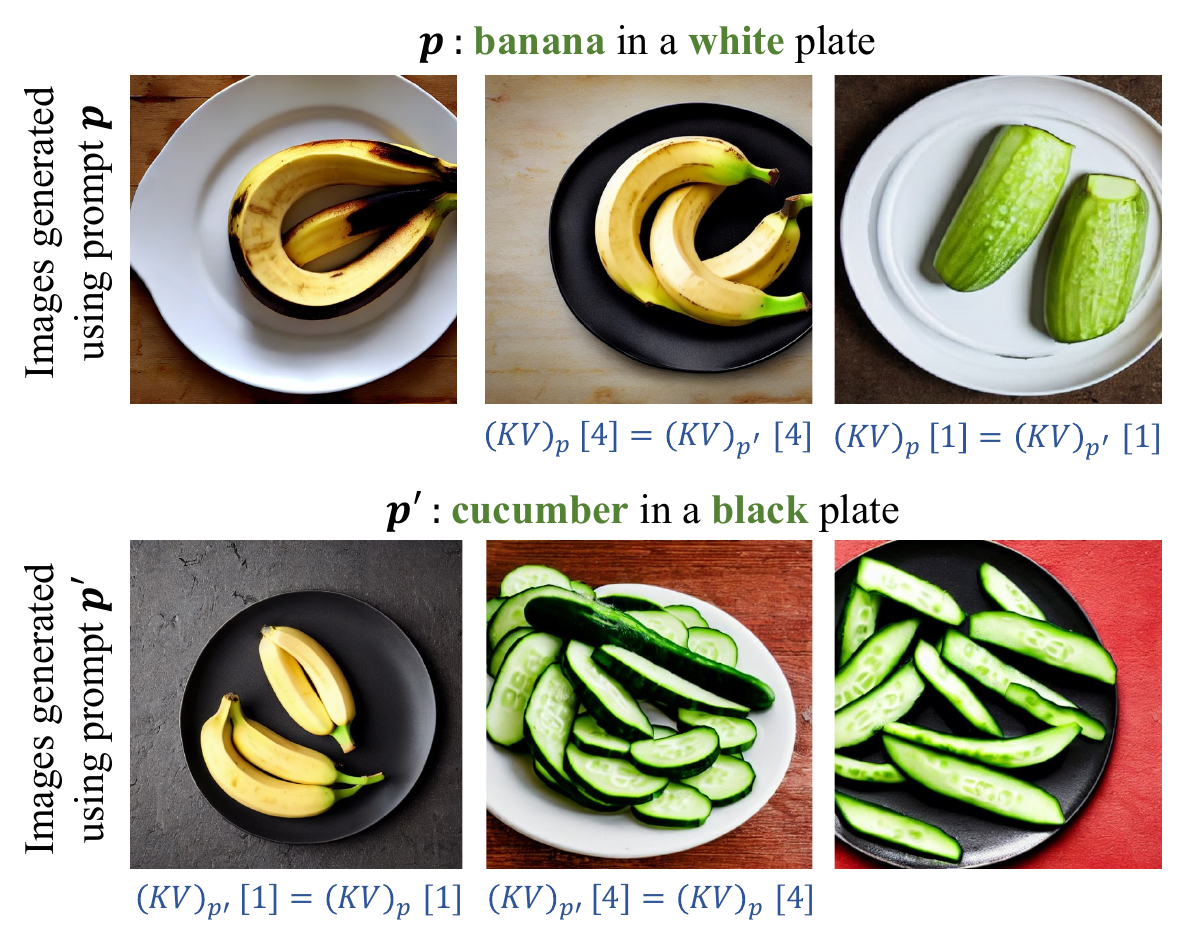}
    \vspace{-8pt}
    \caption{Key-value interactions between two text prompts.}
    % \vspace{-14pt}
    \label{fig:kv_space_analysis2}
\end{figure}

We show more examples in the other columns in Figure \ref{fig:kv_space_analysis1}. For instance, in the first column, we replace the keys and values of \texttt{dog} with a \texttt{cat}, resulting in a cat showing up instead of the dog. We show another set of results in Figure \ref{fig:kv_space_analysis2} that have more semantic interactions between the two text prompts. All the images in the first row are generated using the prompt $p$ \texttt{banana in a white plate}. For the second image, we modify keys and values as $K_p[4] = K_{p'}[4]$, $V_p[4] = V_{p'}[4]$, and hence end up with a banana in a \texttt{black} plate instead (since $p'$ has \texttt{black} at index $4$). Similarly when we modify the keys and values for the second token in $p$, we see \texttt{cucumber} instead of \texttt{banana} in white plate as in the third image in first row. These experiments clearly demonstrate that by manipulating the per-token keys and values, it is possible to control various aspects of the final generation.

\begin{figure*}[h!]
    \centering
    \includegraphics[width = 1.0\linewidth]{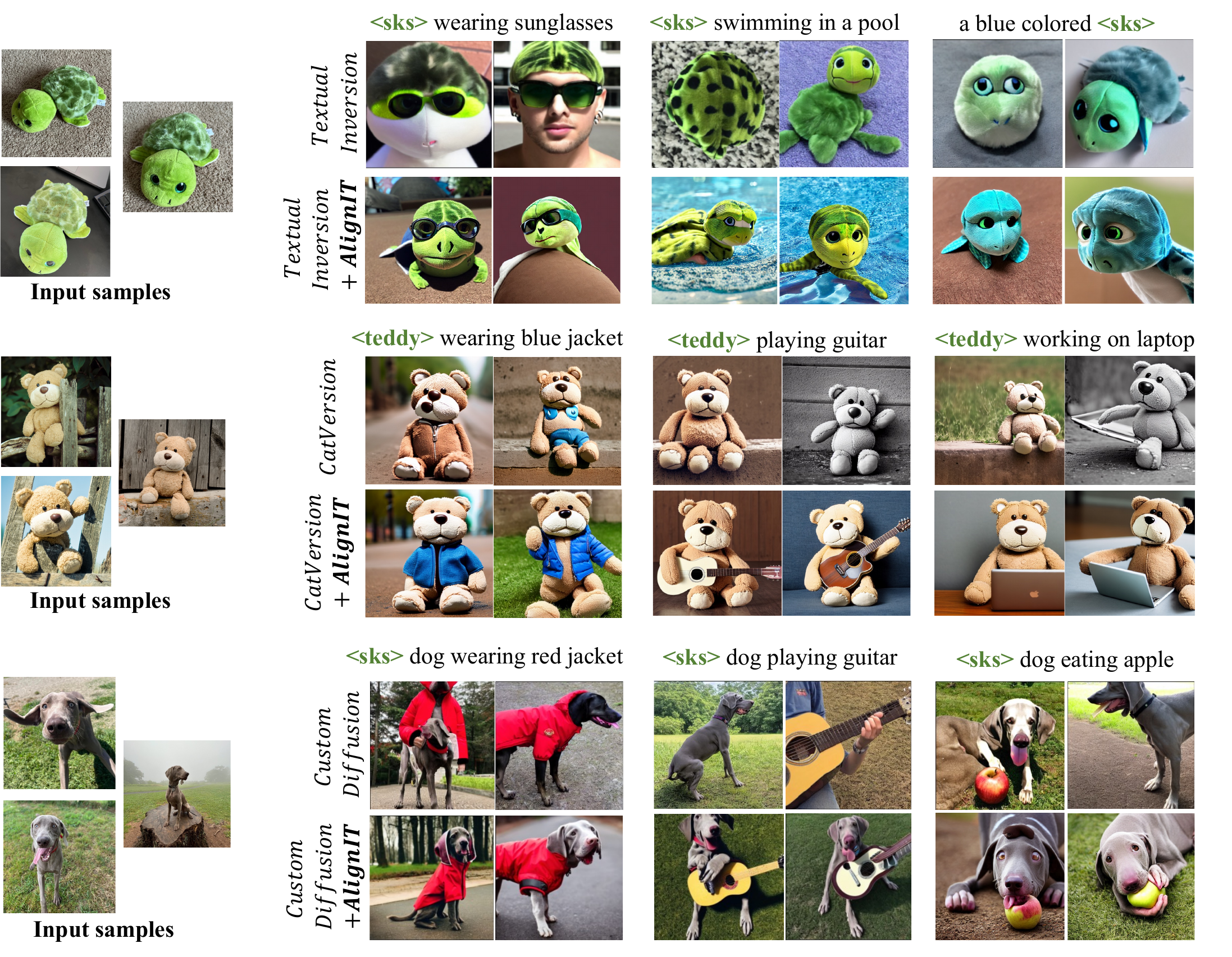}
    \vspace{-8pt}
    \caption{Qualitative comparison with AlignIT plugged into baselines for customized text-to-image generation.}
    \label{fig:results_qual}
\end{figure*}

The experiments and discussion above form the basis for our proposed algorithm \textbf{AlignIT} for customization (presented in Algorithm \ref{alg:AlignIT}). Given a set of reference images for which we want to customize the baseline text-to-image model (e.g., the cat images in the first row of Figure \ref{fig:results_qual}). We assume the existence of a model trained with an existing customization method, e.g., textual inversion \cite{gal2022image}. Now, given a target prompt  (e.g. \texttt{a $<sks>$ dancing in front of times square}) for which we seek to generate an image with the custom concept, we first replace $<sks>$ with a word representing a suitable class belonging to the tokenizer vocabulary to obtain $p$ (\texttt{a cat dancing in front of times square}  in this case). We next use the concept of interest ($<sks>$ here) to construct the dummy prompt $p'$ (`$*$  $<sks>$' in this example) as noted above, and compute the keys $K_{p'}$ and values $V_{p'}$ using the prompt $p'$ with the already available customization model (e.g., textual inversion). These are then used to modify the keys $K_{p}$ and values $V_{p}$ while generating the image using the original prompt of interest $p$ as $K_p[i] = K_{p'}[i]$, $V_p[i] = V_{p'}[i]$ ($i$ = $2$ in this example). Note that while generation with prompt $p$ happens with the baseline stable diffusion model, the knowledge from the reference images (in the form of  keys $K_{p'}$ and values $V_{p'}$) comes from the customization model which is assumed to be already trained for these images.

Finally, since these keys ($K$) and values ($V$) can be computed from any customization method, and are solely responsible for conditioning the image generation process, our proposed method described above can be used on top of any already-trained customization model to improve its performance, which we demonstrate next.

\begin{algorithm}[tb]
    \caption{\textbf{: }AlignIT}
    \label{alg:AlignIT}
    \textbf{Input}: Target prompt $p_t$ having the custom concept token $<sks>$ (belonging to object class $o$) at index $i$\\
    \textbf{Parameter}: Base SD, optimized embeddings/weights from the customized model\\
    \textbf{Output}: Image aligned with prompt $p_t$\\
    \begin{algorithmic}[1]
        \vspace{-10pt}
        \STATE $p \leftarrow p_t$ (Replace $<sks>$ at index $i$ with object class $o$)
        \STATE $p' \leftarrow p_t$ (Replace all except $<sks>$ with $*$)
        \STATE Compute the customized model's $K_p'$ and $V_p'$ for all cross-attention layers with prompt $p'$
        \STATE During generation with Base SD using prompt $p$, in each timestep, do $K_p[i] = K_{p'}[i]$, $V_p[i] = V_{p'}[i]$
    \end{algorithmic}
\end{algorithm}

\begin{figure*}
    \centering
    \includegraphics[width = \linewidth]{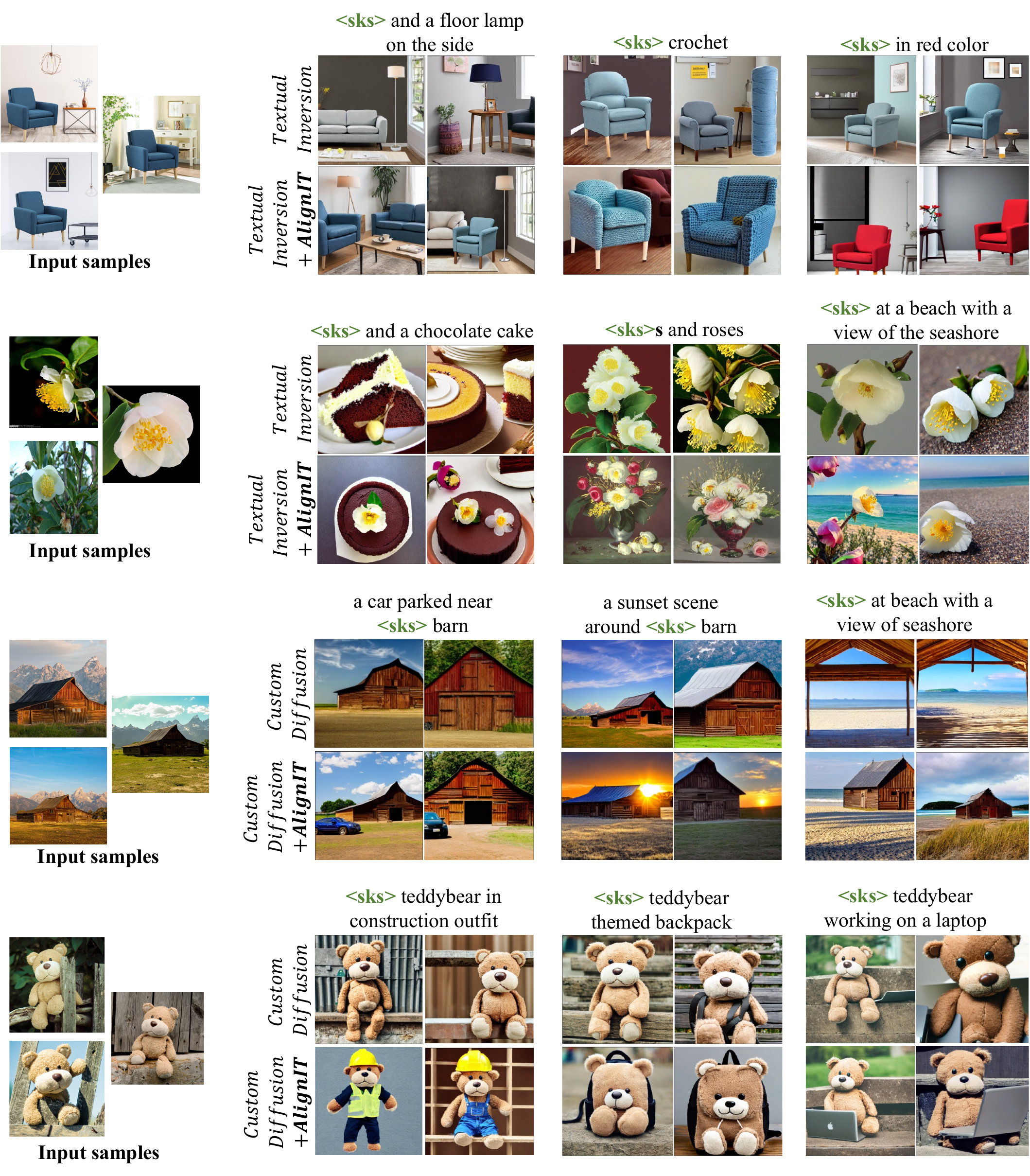}
    \caption{Additional Qualitative Comparison Results with AlignIT plugged into baselines for customized text-to-image generation.}
    \label{fig:results_ti_2}
\end{figure*}

\section{Results}

We conduct extensive qualitative and quantitative experiments using the CustomConcept101 dataset \cite{kumari2023multi} and demonstrate improved customization performance with AlignIT across three different customization methods: Textual Inversion \cite{gal2022image}, Custom Diffusion \cite{kumari2023multi} and CatVersion \cite{zhao2023catversion}.

\textbf{Qualitative Results}
We first begin by discussing our generation outputs. In Figure \ref{fig:results_qual}, we compare AlignIT results when used in conjunction with Textual Inversion \cite{gal2022image}, Custom Diffusion \cite{kumari2023multi} and CatVersion \cite{zhao2023catversion}, and one can note it clearly improves the performance of these methods. For instance, in the first row, Textual Inversion \cite{gal2022image} failed to reconstruct the custom \texttt{tortoise} in the first column, while also missing aspects like \texttt{swimming in a pool} (second column) and \texttt{blue color} (third column). Similarly, in the second row, CatVersion \cite{zhao2023catversion} fails to follow the text prompts fully and misses out concepts like \texttt{jacket}, \texttt{guitar}, and \texttt{laptop}. On the other hand, by using AlignIT, these deficiencies can be alleviated, leading to clearly better-quality results (see the +AlignIT row in each case).

\begin{table*}
    \centering
    \caption{CLIP-based comparisons to quantify AlignIT efficacy. }
    \scalebox{1.0}{
    \begin{tabular}{llll}
        \hline
        Method  & Text Alignment & Image Alignment & Overall \\
        \hline
        Textual Inversion     &   0.67   & 0.83 & 0.75 \\
        Textual Inversion + \textbf{AlignIT} &   \textbf{0.78} \textcolor{blue}{(+16.4\%)}   & \textbf{0.84} \textcolor{blue}{(+1.2\%)} &  \textbf{0.81} \textcolor{blue}{(+8.0\%)}\\
        \hdashline[1.5pt/3pt]
        CatVersion     &   0.73   & 0.82 &  0.77\\
        CatVersion + \textbf{AlignIT} &  \textbf{0.79} \textcolor{blue}{(+8.2\%)}  & \textbf{0.84} \textcolor{blue}{(+2.4\%)} &  \textbf{0.81} \textcolor{blue}{(+5.2\%)}\\
        \hdashline[1.5pt/3pt]
        Custom Diffusion &   0.77   & 0.82 & 0.79 \\
        Custom Diffusion + \textbf{AlignIT} & \textbf{0.81} \textcolor{blue}{(+5.2\%)}     & \textbf{0.83} \textcolor{blue}{(+1.2\%)}&  \textbf{0.82} \textcolor{blue}{(+3.8\%)}\\
        \hline
    \end{tabular}
    }
    \vspace{6pt}
    \label{tab:quant_eval}
\end{table*}

\begin{table*}
    \centering
    \caption{Results from a user survey with 24 respondents. }
    \scalebox{1.0}{
    \begin{tabular}{llll}
        \hline
        Method  & Text Alignment & Image Alignment & Overall \\
        \hline
        Textual Inversion     &   4.3\%   & 12.4\% &  7.3\% \\
        Textual Inversion + \textbf{AlignIT} &  \textbf{95.7}\%    & \textbf{87.6}\% &  \textbf{91.6}\%\\
        \hdashline[1.5pt/3pt]
        Custom Diffusion &  7.1\% & 11.8\% & 9.15\% \\
        Custom Diffusion + \textbf{AlignIT} & \textbf{92.9}\%  & \textbf{88.2}\%  &  \textbf{90.5}\% \\
        \hline
    \end{tabular}
    }
    \vspace{6pt}
    \label{tab:user_study}
\end{table*}

\textbf{Quantitative Results}
For customization methods, evaluating both the ability to replicate the custom concept, and the ability to modify the custom concept using textual prompts is important. We follow the existing protocol \cite{zhao2023catversion,kumari2023multi} and quantify performance using CLIP-based distances. The CustomConcept101 dataset has a set of 20 curated prompts for each concept. As in prior work \cite{kumari2023multi}, we generate $50$ images with randomly selected seeds for each prompt, giving us $1$K generated images for each concept. We measure CLIP text alignment score by computing the average similarity between the text prompt and the generated images for each prompt and concept, thereby evaluating the ability to modify the custom concept using textual prompts. We next follow CatVersion \cite{zhao2023catversion} and adjust the CLIP image alignment score to better focus on the similarity between the concept of interest and corresponding reference images to evaluate the concept reconstruction quality. We do this by computing masks for the concept of interest in the generated images and measuring similarities by discarding the pixels that do not belong to the concept of interest. We also report the geometric mean of the image and text alignment scores to get an estimate of the overall performance. Table \ref{tab:quant_eval} summarizes these results where much higher CLIP similarities are indicative of the improved customization effect of the generated results with our method when compared to the baselines. One can note that AlignIT dramatically improves the CLIP text alignment scores while maintaining high image alignment.

\textbf{User Study} 
Finally, we conduct a user study with generated images and evaluate the mean preference of AlignIT plugged with two baselines. Each participant is asked a set of 20 questions where we ask them to select either the image (among pair of images, each belonging to a baseline and AlignIT applied on top of baseline) that best aligns with the prompt or the one that best reconstructs the concept of interest given a reference image. From Table \ref{tab:user_study}, one can clearly note that the users prefer the case where AlignIT is applied on top of the baselines in all of text-guided alignment, reconstruction fidelity and overall customization effect.

\section{Conclusion}
In this work, we noticed that existing customization techniques fail to generate images that fully align with user's intent. We first discuss reasons behind failure of existing works. We demonstrated that existing methods (during inference) undesirably end up affecting the keys and values for tokens other than the custom concept of interest as well, thereby leading to misaligned images. To address these issues, we proposed an algorithm called AlignIT that can be plugged into any of these existing customization methods and fix these issues directly at test-time. We conducted extensive qualitative experiments on the CustomConcept101 dataset and demonstrated that the images generated after plugging AlignIT with the existing baselines are substantially more aligned with the input prompts, while also retaining the reconstruction quality of the concept of interest. Further, we also quantified our improvements with existing protocols and a user survey that clearly showed the efficacy of AlignIT.

{\small
\bibliographystyle{ieee_fullname}
\bibliography{egbib}
}

\end{document}